\pdfoutput=1
\documentclass[11pt]{article}

\usepackage{fullpage,times}
\usepackage{parskip}
\usepackage{tikz}
\usetikzlibrary{bayesnet}
\usetikzlibrary{arrows}
\usepackage{color}
\usepackage{graphicx}
\usepackage{bm}
\usepackage[inline]{enumitem}

\usepackage{amsmath, amsthm, amssymb, amsfonts, mathtools, graphicx, enumerate}
\usepackage{adjustbox}
\usepackage[utf8]{inputenc} 
\usepackage[T1]{fontenc}    
\usepackage{url}            
\usepackage{booktabs}
\usepackage{nicefrac}       
\usepackage{microtype}      
\usepackage{xspace}
\usepackage{algorithm,algorithmic}
\usepackage{color}
\usepackage{enumitem}
\usepackage{comment}
\usepackage{bm}
\usepackage{apptools}
\usepackage[page, header]{appendix}
\usepackage{titletoc}
\usepackage{physics}
\usepackage{array}

\usepackage{subcaption}

\usepackage[scaled=.9]{helvet}

\usepackage{url}
\usepackage{caption}
\usepackage{natbib}
\bibpunct{(}{)}{;}{a}{,}{,}

\usepackage{amsthm}   
\newtheorem{thm}{Theorem}[section]

\newtheorem{mydef}[thm]{Definition}

\begingroup
    \makeatletter
    \@for\theoremstyle:=definition,remark,plain\do{%
        \expandafter\g@addto@macro\csname th@\theoremstyle\endcsname{%
            \addtolength\thm@preskip\parskip
            }%
        }
\endgroup

\newcommand{\myempty}[1]{}

\DeclareMathOperator*{\argmin}{argmin}

\newcommand{\yrm}[1]{}

\newcommand{\qiangremoved}[1]{}

 \def\bb#1\ee{\begin{align*}#1\end{align*}}

 \def\bba#1\eea{\begin{align}#1\end{align}}

\usepackage{lipsum}  
\makeatletter
\newcommand{\printfnsymbol}[1]{%
  \textsuperscript{\@fnsymbol{#1}}%
}
\makeatother

\newif\ifshowcomments
\showcommentstrue
\showcommentsfalse

\title{Learning a Shield from Catastrophic Action Effects: \\ Never Repeat the Same Mistake}
\author{%
	Shahaf S. Shperberg\\
    ~University of Texas at Austin\\
	\texttt{shperbsh@cs.utexas.edu}
	\and
	Bo Liu\\
	~University of Texas at Austin\\
	\texttt{bliu@cs.utexas.edu}
	\and 
	Peter Stone\\
	University of Texas at Austin, Sony AI\\
	\texttt{pstone@cs.utexas.edu} \\
}

\date{}

\begin{document}
\maketitle



\begin{abstract}

    Agents that operate in an unknown environment are bound to make mistakes while learning, including, at least occasionally, some that lead to catastrophic consequences. When humans make catastrophic mistakes, they are expected to learn never to repeat them, such as a toddler who touches a hot stove and immediately learns never to do so again.  In this work we consider a novel class of POMDPs, called POMDP with Catastrophic Actions (POMDP-CA) in which pairs of states and actions are labeled as catastrophic. Agents that act in a POMDP-CA do not have a priori knowledge about which (state, action) pairs are catastrophic, thus they are sure to make mistakes when trying to learn any meaningful policy. Rather, their aim is to maximize reward while never repeating mistakes. 
    
    As a first step of avoiding mistake repetition, we leverage the concept of a \emph{shield} which prevents agents from executing specific actions from specific states.
    In particular, we store catastrophic mistakes (unsafe pairs of states and actions) that agents make in a database. Agents are then forbidden to pick actions that appear in the database. This approach is especially useful in a continual learning setting, where groups of agents perform a variety of tasks over time in the same underlying environment. In this setting, a task-agnostic shield can be constructed in a way that stores mistakes made by any agent, such that once one agent in a group makes a mistake the entire group learns to never repeat that mistake. This paper introduces a variant of the PPO algorithm that utilizes this shield, called ShieldPPO, and empirically evaluates it
    in a controlled environment.
    Results indicate that ShieldPPO outperforms PPO, as well as baseline methods from the safe reinforcement learning literature, in a range of settings.
\end{abstract}

\section{Introduction}

Mistakes are an inevitable part of any learning process.  However, not
all mistakes are created equal.  On the one hand, any suboptimal action,
no matter how benign, can be considered a mistake.  Such slightly
suboptimal actions may be usefully repeated on the path towards
perfection.  On the other hand, the consequences of some mistakes are so
dire that they must never be repeated.  
Consider, for example, an autonomous car that is learning to optimize
its performance in terms of some combination of travel time, fuel
efficiency, passenger comfort, and safety. We might tolerate, even
expect, the car to repeatedly experiment with suboptimal rates of
acceleration and deceleration, cruising velocities, lane placements,
etc. while gradually improving performance. However, if the car ever
takes a catastrophic action such as rear-ending the car in front of it,
we would like that mistake never to be repeated, ideally neither by that
car, nor by any other. 

Conceptually, different mistakes can be associated with different events that occur in the environment. For example, the first time a dog jumps in front of a vehicle, the vehicle is faced with a new risk of collision. While we prefer that the vehicle never makes any mistakes, even human drivers are not perfect.  However, unlike human drivers, if we can ensure that once a car makes a certain type of mistake then it will never repeat it, then the vehicle (and the rest of its fleet) will become monotonically safer over time.

Autonomous agent learning is commonly modeled as a Markov Decision Process (MDP) or when the environment cannot be fully observed, its Partially Observable variant (POMDP). The learning process is typically evaluated based on whether the learned policy is
eventually optimal~\citep{watkins1989learning}, based on how many
suboptimal actions are executed while learning~\citep{brafman2002r}, or based on
cumulative reward relative to that of the optimal policy~\citep{lai1985asymptotically}.  However, none of these metrics
captures the notion that catastrophic actions must never be repeated.

This paper thus introduces\footnote{A related formulation has been considered in contemporary, unpublished work \citep{he2021androids}.} the concept of a POMDP with Catastrophic
Actions (POMDP-CA) that designates certain (state, action) pairs as
\emph{catastrophic}. Of course, if such pairs can be identified ahead of
time, they should simply be removed from the POMDP so that they are never
executed. However, inevitably, not all catastrophic actions can be
foreseen at design time.  Rather, they must be experienced once, but
then can be stored in a monotonically growing database 
of (state, action) pairs 
never to be executed again.  All future policies of the
learning agent, and in a multiagent setting of any other agent, can then
be validated against this database \emph{before deployment} to ensure
that no (state, action) pairs known to be catastrophic will ever be
executed.

With this motivation in mind, this paper considers the problem of
maximizing performance in a \emph{sequence} of POMDP-CAs subject to the constraint that no catastrophic actions may be executed more than once. The POMDP-CAs all share very similar environments, except that each may contain a small number of idiosyncratic states and transitions that represent rarely occurring events (such as the dog jumping in front of the car). Some events are expected to happen relatively frequently, while other events are not likely to ever repeat~\citep{Waymo}.
To capture that notion, we assume that each POMDP-CA in the sequence is drawn from an underlying long-tailed distribution.

We take a first
step towards addressing this problem by introducing a straightforward
extension to the Proximal Policy Optimization (PPO) algorithm~\citep{schulman2017proximal}
called ShieldPPO based on the (admittedly limiting) assumptions that the
environment is discrete, the number of relevant catastrophic (state,
action) pairs is small enough to store in memory, and agents can
identify such catastrophic actions as they occur.  As the name suggests,
ShieldPPO leverages the concept of a \emph{shield}~\citep{DBLP:conf/aaai/AlshiekhBEKNT18} to constrain agents' actions. However, unlike most past work, the shield is
learned online.

In empirical evaluations in a toy domain, we find that ShieldPPO
outperforms PPO, as well as a range of \emph{safe-RL} algorithms, both
in terms of number of catastrophic actions executed and mean episodic
return.  The advantages of ShieldPPO are particularly pronounced in the
case of multiple learning agents that can share experience, and in a
continual learning setting with multiple tasks in the same underlying
environment such that each task shares the same set of catastrophic
actions.

As the introduction of POMDP-CAs, this paper focuses on a minimal
representative setting for the purpose of in-depth analysis.
Despite the simplifying assumptions and correspondingly constrained
evaluation setting, this initial treatment of POMDP-CAs introduces an
important, previously unexplored line of research, rife with
possibilities for future extensions.

\section{Related Work} \label{sec:related}

Safe reinforcement learning (safe RL) studies how to maximize environment return while being safe. There are three major perspectives on achieving safety. The constrained optimization perspective assumes that besides the environment reward, the agent receives an additional cost per step~\citep{kadota2006discounted,moldovan2012safe}. Therefore, to behave safely, the agent aims to maximize the environment return subject to the discounted cumulative cost being below some given threshold. The constrained optimization problem is usually formulated as a Constrained Markov Decision Process (CMDP)~\citep{altman1999constrained}. To solve the constrained optimization problem, augmented Lagrangian-based methods~\citep{liu2020ipo}, trust-region methods~\citep{achiam2017constrained} and Lyapunov-based methods~\citep{chow2018lyapunov} are proposed. These methods do not address the problem of mistake repetition, either agents repeat mistakes or the threshold is low enough such that agents avoid making mistakes at all, even if it means achieving low rewards (e.g., by not moving).

The second perspective aims to minimize the risk of being unsafe. This means the agent considers the future return as a distribution rather than an expectation (e.g., the agent not only optimizes the return but also tries to reduce the uncertainty of achieving that return). As a result, the agent avoids performing risky but highly rewarding actions. To balance the risk and return, various criteria can be used, including %
Value-at-Risk (VaR)~\citep{mausser1999beyond}, coefficient of determination (a.k.a $r$-squared), and the Sharpe value~\citep{di2012policy}. For a more comprehensive review, 
refer to a prior survey of safe reinforcement learning methods~\citep{garcia2015comprehensive}. Similarly to CMDP-based methods, risk-based methods do not guarantee that mistakes are never repeated, unless the risk threshold is set to an extreme, resulting in a degenerate policy that obtains low rewards.

The final perspective is safety via shielding as described in Section~\ref{sec:backgraound:shielding}. \cite{DBLP:conf/aaai/AlshiekhBEKNT18} have considered the case in which a safety constraint $\phi$ is given as an input to the agent. %
Using this constraint, they developed a reactive system that acts as a shield and ensures that the given constraint is never violated. Similar approaches have been applied for multi-agent reinforcement learning~\citep{DBLP:conf/atal/Elsayed-AlyBAET21}, and for robots with known dynamics~\citep{DBLP:conf/amcc/Bastani21}. Another notable work introduces a method called Bounded-prescience shield (BPS) for shielding via an emulator that verifies which actions of the current state are safe with respect to a given safety criterion by looking ahead into future states up to a bounded horizon \citep{giacobbe2021shielding}. Finally, the work of \cite{he2021androids} relaxes the need for emulator of BPS by learning a latent model and doing simulation in the latent model instead of in an emulator. They assume that there exists a function $L_\phi$ that indicates which states are considered to be safe (as opposed to state and action pairs, as in this work) and that whenever the agent reaches a state $s$, $L_\phi(s)$ is observable. To the best of our knowledge, that contemporary, unpublished work is the only other work which considers online construction of a shield.

\section{Background and Notation} \label{sec:backgraound}
This section provides background on reinforcement learning, Partially Observed Markov Decision Processes, and safety in reinforcement learning using a shield.%

\subsection{Partially Observable Markov Decision Process}
Reinforcement learning (RL) is often formulated as a Partially Observable Markov Decision Process (POMDP). A POMDP is a tuple $M = (S, \mathcal{A}, T, \gamma, R, \Omega, \mathcal{O})$. Here, $S$ and $\mathcal{A}$ are the state and action spaces, respectively. $T$ is the transition dynamics, i.e. at state $s$, taking action $a$ will have $T(s' \mid s, a)$ probability of transitioning to $s'$. $\gamma$ is the discount factor and $R : S \times \mathcal{A} \rightarrow \mathbb{R}$ is the reward function, such that $R(s,a)$ is the reward obtained by executing action $a$ from state $s$. Finally, $\Omega$ is the set of observations, and $\mathcal{O}: S \times A \times \Omega \rightarrow \mathbb{R}$ is a conditional observation probability function. In a POMDP, the subsequent state $s'$ resulting from executing an action $a$ from a state $s$ is not observable to the agent. Instead, it receives an observation $o \in \Omega$ with probability $\mathcal{O}(o|s',a)$. 
The agent's objective is to learn a policy $\pi_\theta: S \rightarrow \mathcal{A}$ parameterized by $\theta$, that optimizes the discounted cumulative reward
${\mathbb{E}}_{(s_t, a_t)\sim \pi} \big[\sum_{t=0}^\infty \gamma^t R(s_t, a_t)\big].$

\subsection{Safe RL via Shielding} \label{sec:backgraound:shielding}
Safety is an important criterion in real-world applications of RL. A simple but effective approach for safe RL is to ensure that no ``unsafe" actions are taken. To this end, the shielding method was proposed, in which unsafe actions are directly masked out~\citep{DBLP:conf/aaai/AlshiekhBEKNT18}. Formally, a shield is a binary function $\mathcal{S}: S \times \mathcal{A} \rightarrow \{0, 1\}$. Here, a state and action pair $(s,a)$ is deemed as safe with respect to $\mathcal{S}$ if $\mathcal{S}(s,a) = 1$ and unsafe otherwise.
Given the agent's policy $\pi_\theta$, a shielded policy with respect to a shield $\mathcal{S}$ is defined as:
\begin{equation}
    \pi_{\theta}^{\mathcal{S}} (a|s) = 
    \begin{cases}
    \frac{1}{Z}~\pi_\theta(a | s)~\mathcal{S}(s, a) & \sum_{a'} \mathcal{S}(s,a') \geq 1, \\
    \pi_\text{default}(a|s) & \text{otherwise},
    \end{cases}
\end{equation}
where $Z = \sum_{a'} \pi_\theta(a'|s)\mathcal{S}(s,a')$ is the normalization term and $\pi_\text{default}$ is some default policy in case no actions are safe from state $s$ according to $\mathcal{S}$. The above transformation is referred to as \emph{applying} the shield. A shield can be applied either when deploying a fully-trained policy or during training. One advantage of applying a shield during training is more efficient experience collection, as the agent will avoid taking actions which are known to be unsafe. Note that this method of applying the shield assumes that the shield is accurate, since if there are errors in the shield (e.g., as a result of approximation), optimal (safe) actions can potentially be discarded.%

\section{Problem Definition} \label{sec:problemDef}
In this work, we extend the definition of POMDPs to incorporate \emph{catastrophic actions}, introducing a new type of problem called a POMDP with Catastrophic Actions (POMDP-CA).  A POMDP-CA includes an additional 
\emph{ground truth} safety labeling function $L_\phi: S \times \mathcal{A} \rightarrow \{0, 1\}$, such that $L_\phi(s,a) = 1$ if and only if taking action $a$ at state $s$ is safe. 
Using $L_\phi$, we define catastrophic state-action pairs.
\begin{mydef}[Catastrophic State-Action Pairs]
Given a ground truth safety labeling function $L_\phi$, the set of catastrophic state-action pairs is defined as:
\begin{equation}
    \mathbb{C} = \bigg\{(s,a)~\bigg|~ L_\phi(s,a) = 0,~\text{where}~s \in S, a \in \mathcal{A}~\bigg\}.
    \label{eq:catastrophic-sa}
\end{equation}
\end{mydef}

The problem we consider is a sequence of $K$ POMDP-CAs, $\{M_i = (S_i, \mathcal{A}_i, T_i, R_i, \gamma_i, \Omega_i, \mathcal{O}_i, L_\phi)\}_{i=1}^K$, such that all POMDP-CAs share the same safety function $L_\phi : S \times \mathcal{A}$, and that for every $i$, $S_i \subseteq S$ and $\mathcal{A}_i \subseteq \mathcal{A}$.
As the agent progresses to the $k$-th POMDP-CA, its objective is to find a policy $\pi$ that optimizes the discounted cumulative reward on all $\{M_i\}_{i \leq k}$:
\begin{equation}
    J(\pi) = \sum_{i=1}^k \underset{(s^i_t,a^i_t) \sim \pi, M_i}{\mathbb{E}}~\bigg[ 
      \sum_{t=0}^{\infty} \gamma_i^t R_i(s_t, a_t)
      \bigg].
\end{equation}
Essentially, the agent learns a \emph{single} policy $\pi : S \rightarrow \mathcal{A}$ that is being projected to $S_i$ for each POMDP-CA in the sequence.

Conceptually, all POMDP-CAs share very similar environments, with the exception of a small number of idiosyncratic states and transitions that each POMDP-CA may contain. These idiosyncratic states and transitions represent
rarely occurring events (e.g., a dog jumping in front of
the car). To capture the notion that such idiosyncratic
states and transitions occur with different frequencies, each POMDP-CA in the sequence is assumed to be drawn from an underlying long-tailed distribution $\mathbb{M}$, such that for every $i$, $M_i \sim \mathbb{M}$.

\section{Learning a Shield Online} \label{sec:method}
If the set of catastrophic actions from each state ($L_\phi$) is known in advance, then one can simply define the shield such that the agent is prohibited from taking such actions.  However, in this paper we are interested in the case where $L_\phi$ is \emph{not} known in advance.  Thus, the shield must be learned in an online manner by
discovering catastrophic pairs $(s,a)\in \mathbb{C}$.

On the surface, the concept of this section is very straightforward: don't take actions that have been found to be catastrophic. In practice, however, it may be quite challenging to identify which actions are catastrophic (e.g., which action, from which state, was the root cause of the car's accident). In this work, we sidestep this challenge by making three main assumptions: \begin{enumerate*}[label=(\roman*)]
    \item the environments are discrete (Section~\ref{sec:non-parametric}),
    \item the amount of common catastrophic-pairs is small enough to store in memory (Section~\ref{sec:non-parametric}), and
    \item agents can observe $L_\phi$, i.e., identify catastrophic mistakes as they occur (Section~\ref{sec:identify}).
\end{enumerate*}
Clearly, these assumptions do not hold in all domains, but we make them for the purpose of an initial analysis that can inform further studies with relaxed assumptions.

\subsection{Non-parametric (Tabular) Shield} \label{sec:non-parametric}
The most intuitive method to learn a shield by observations is to store every catastrophic pair in a table $\mathcal{T} = \{(s, a)\}$ (e.g., a dictionary). In this way, the shield $\mathcal{S}_\mathcal{T}$ can be defined as:
\begin{equation}
\mathcal{S}_\mathcal{T}(s,a) = \begin{cases}
    1 & (s,a) \notin \mathcal{T}\\
    0 & \text{otherwise}.
\end{cases}
\end{equation}

While this approach is very simple, it has some appealing advantages. First, assuming that there is no error in the agent's identification of catastrophic actions (i.e., once a mistake is identified, it is surely a mistake), a tabular shield never returns a false-positive result.
Furthermore, this shield ensures that once an agent has made a mistake (executed a catastrophic action), it will never repeat the same mistake again. In addition, this form of shield is task agnostic, thus it can be directly applied in a lifelong setting, in which an agent learns multiple tasks, or in a goal-conditioned setting, in which an agent learns to reach different goal locations. Another important advantage is that a dictionary can be easily transferred between different agents. Moreover, sharing a tabular shield ensures that a mistake made by \emph{one} of the agents will never be repeated by \emph{any} agents. Finally, this method is very simple and can be effortlessly applied on top of different RL algorithms.

Nonetheless, there are also some drawbacks to using a tabular shield. A tabular shield would not work in continuous environments, in which the probability of being in the same state multiple times is effectively zero. Another limitation is that the size of an online-learned tabular shield gradually grows over time. Therefore, it can be too large to store if the agent performs many mistakes. Furthermore, the query time 
increases with the table size. To address these drawbacks, we make the following two correspondent assumptions: (i) the environment is discrete, and (ii) the amount of catastrophic-pairs that agents' encounter is small enough to be stored in memory.

There are several ways to address the memory limitations. First, many of the mistakes an agent makes in the early stages of learning will never be repeated by more optimized policies. Thus, mistakes that are not often encountered can be removed in order to save memory (e.g., in a least-recently-used manner). Another way to improve runtime and to save memory is to implement the dictionary using monotone minimal perfect hashing and to efficiently encode the state-action pairs~\citep{navarro2016compact}. An alternative to a dictionary is a Bloom filter~\citep{Bloom70}. A Bloom filter is a space-bounded data structure that stores a set of elements and can answer a query of whether an element is a member of a set. Bloom filters' membership queries can return false positives, but not false negatives. Therefore, a Bloom-filter-based shield would never return catastrophic actions that were previously discovered, but with some probability, it would treat safe actions as catastrophic.
Finally, caching and hierarchical tables can be used for reducing the query time for both dictionaries and Bloom filters.

\subsection{Parametric Shield}
An alternative to learning a tabular shield is to learn a parametric shield
$\mathcal{S}_\theta$ based on catastrophic pairs encountered by the agent.
A simple way of learning a shield is by doing binary prediction (e.g., logistic regression):
\begin{equation}
    \theta^* = \argmin_\theta
      \begin{pmatrix}
    \mathbb{E}_{(s,a) \in \mathcal{T}^C} \big[\log (\mathcal{S}_\theta(s,a))\big]+\\
     \mathbb{E}_{(s,a) \in \mathcal{T}}\big[\log (1 - \mathcal{S}_\theta(s,a))\big]
  \end{pmatrix}
\end{equation}

A benefit of a parametric shield in terms of memory and runtime is that the size of the function approximator is constant, as is the query time. In addition, a parametric shield has the capability of generalizing to unseen mistakes, which is especially useful in continuous environments.
Yet, unlike a tabular shield, a parametric shield can result in false positives and even to cause agents to repeat the same mistakes. A possible compromise between the two approaches is to use a \emph{hybrid} shield, e.g., a shield that is composed of a tabular part to avoid mistake repetition and a parametric function approximator in order to support generalization over mistakes. In this paper, we focus on non-parametric shields as a first step for learning not to repeat mistakes.

\subsection{Identifying Mistakes and Their Triggers} \label{sec:identify}
A key challenge in learning a shield online is identifying when an agent has made a mistake. In principle, any suboptimal action can be treated as a mistake. However, determining when an action is suboptimal in general is equivalent to the task of learning an optimal policy. Therefore, we aim only to avoid repeating catastrophic mistakes. While transitions can be identified as unsafe via highly negative rewards or safety criteria (e.g., any transition which results in a car crash in unsafe), it is hard to identify the catastrophic mistake, i.e., which action from which state was the cause of the incident. For example, consider the following simple part of an MDP:
\begin{equation}
  s_1 \xrightarrow{1,2} s_2 \xrightarrow{1,2} s_3 \dots \xrightarrow{1,2} s_n
\end{equation}
Here the action space is $\mathcal{A} = \{1, 2\}$ and reaching $s_n$ is unsafe. Even if an agent could detect that transitions from $s_{n-1}$ to $s_n$ are unsafe, the actual catastrophic mistakes are actions that leads to $s_1$, as by then $s_n$ is unavoidable. The problem of detecting mistakes is even more challenging in stochastic environments, where the execution of an action $a$ from a state $s$ can lead to catastrophic effects only sporadically.

Therefore, in this work we further assume that: (iii) $L_\phi$ is exposed to the agent via feedback. For instance, when the agent takes action $a$ at a state $s$, it receives not only a reward $r \sim R(s,a)$ but also the safety label $u = L_\phi(s, a)$. This strong assumption can be justified in two ways. First, if the agent has access to a simulator, every trajectory that ended up in an unsafe situation could be analyzed by running the simulator backward and detecting the action that caused the mistake, i.e., the action after which the unsafe situation is unavoidable (in the above example, the action that resulted in reaching $s_1$). Alternatively, if the rate of mistakes is sufficiently low, the cause of a mistake could be identified by domain experts; this is already being done in the case of aviation accidents~\citep{FAA} or car accidents that involve autonomous vehicles~\citep{sinha2021crash}. Thus, even if $L_\phi$ is not directly observable by the agent, there are cases in which such an observation can be given to the agent ex post facto, after it has made a mistake, via an external analysis process. Ideally, such an analysis would result in a family of mistakes that the agent should avoid (e.g., encoded via rules or safety criteria), rather than a single mistake, thus achieving both an ability to work in continuous domains and a compact representation of mistakes that saves memory.

\section{Empirical Evaluation} \label{sec:evaluation}

To study the effect of the tabular shield, we apply it to the Proximal Policy Optimization (PPO) algorithm~\citep{schulman2017proximal}, resulting in a variant called ShieldPPO.
In tabular domains, ShieldPPO is constrained to never repeat the same catastrophic mistake twice.  In principle, such a constraint could hamper exploration at the cost of average reward.  Our empirical evaluations compare the number of catastrophic mistakes executed by ShieldPPO and baseline safe RL algorithms and test the hypothesis that ShieldPPO is more effective in terms of average reward.  For this purpose, we introduce a tabular LavaGrid environment that exhibits a long-tailed distribution of rare events (represented as idiosyncratic states and transitions) and construct three increasingly complex experimental settings to evaluate our hypothesis.  Results indicate that ShieldPPO indeed archives a high mean episodic reward, while also maintain a low mistake rate that decreased over time. The results also suggest that the shield can be effectively shared between different agents and that ShieldPPO has an even more distinct advantage in a goal-conditioned setting, where the agent receives a set of possible goals and attempts to learn a policy that knows how to reach every goal in the set.

\subsection{Baselines}

The first algorithm that acts as a natural baseline in our experiments is the basic version of PPO, which does not use a shield. In addition, we consider other algorithms from the safe RL literature as baselines. The first two safe RL baseline algorithms were designed for solving CMDPs: Constrained policy optimization (CPO, \cite{achiam2017constrained}), which enforces constraints throughout training by solving trust region optimization problems at each policy update, and PPO-Lagrangian~\cite{ray2019benchmarking}, a variant of PPO that enforces constraints by using adaptive penalty coefficients. Our final baseline is Worst-Case Soft Actor Critic (WCSAC, \cite{yang2021WCSAC}), a state-of-the-art Value-at-Risk-based method that achieves risk control by extending the Soft Actor Critic algorithm~\citep{haarnoja2018soft} with a safety critic.

\subsection{Experimental Settings}
Our experiments are conducted on three different settings. First, a single-agent setting in which one agent is faced with a sequence of LavaGrid instances, as described below in Section~\ref{sec:domain}. The next experiment assumes that there are multiple agents which simultaneously face LavaGrid environments. Here, the agents do not encounter the same sequence of environments, but rather each agent faces its own sequences of POMDP-CAs, where all environments are drawn from the same distribution. This setting is meant to represent, for example, a fleet of autonomous cars, each driving in the same city every day. In this experiment we evaluate different levels of shield sharing which correspond to three variants of PPO, one that does not use a shield, one in which each agent has its own individual shield, and lastly, a variant in which all agents share the same shield. Finally, we test ShieldPPO and the different baselines in a goal-condition setting, where the agent needs to learn a single policy that can reach different goals. 

\subsection{The LavaGrid Domain} \label{sec:domain}
\begin{figure*}[tbh]
     \centering
      \begin{subfigure}[t]{0.48\textwidth}
         \centering
         \includegraphics[width=\textwidth]{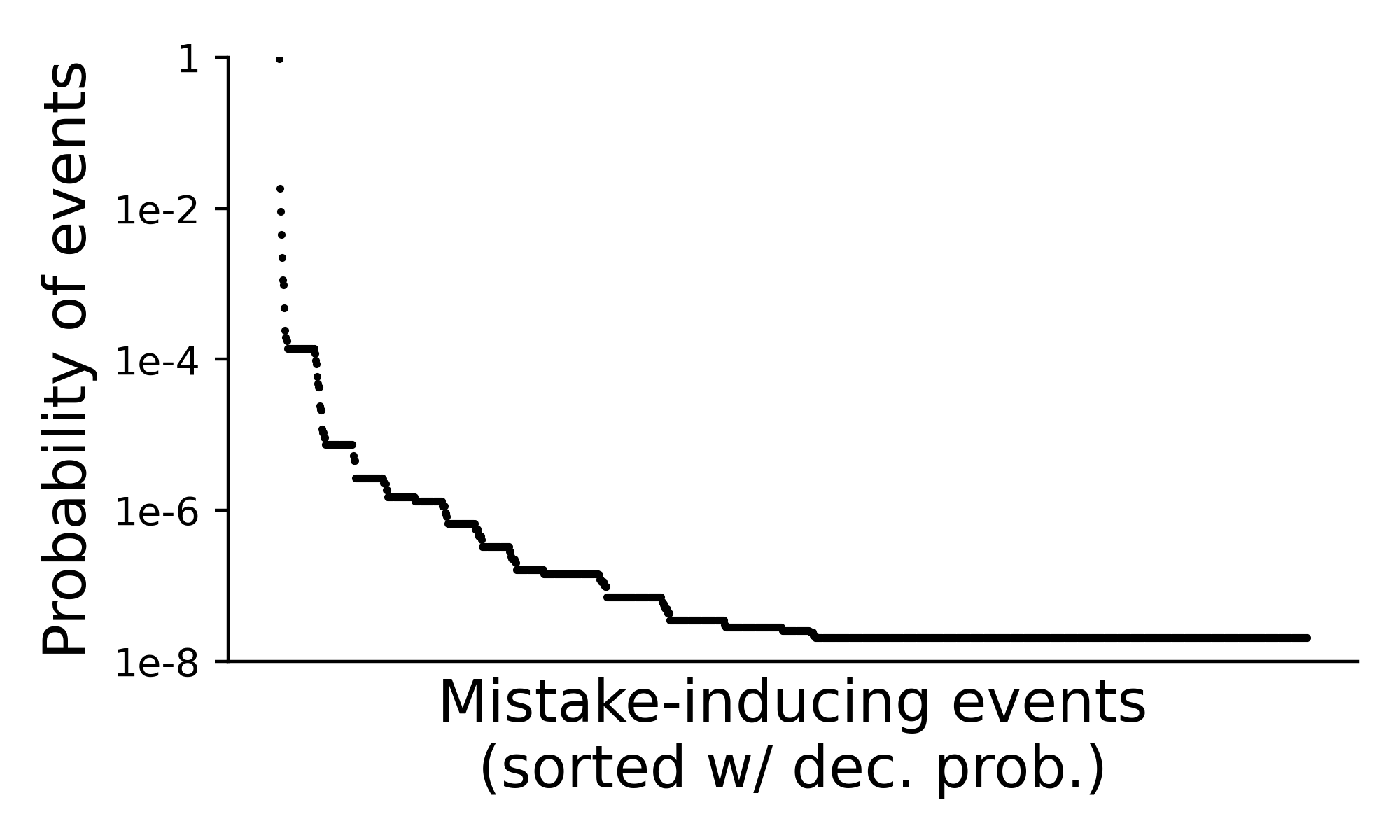}
         \label{fig:lavaA}
     \hfill
     \end{subfigure}
     \begin{subfigure}[t]{0.45\textwidth}
         \centering
         \includegraphics[width=\textwidth]{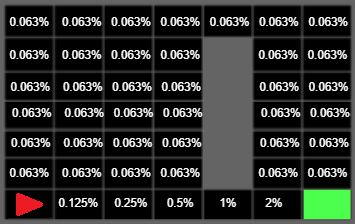}
         \label{fig:lavaB}
     \end{subfigure}
        \caption{The LavaGrid domain; \textbf{left:} distribution of potential mistake-inducing events (log scale), \textbf{right:} distribution of events}    
        \label{fig:LavaGrid}
        
\end{figure*}

In order to have a controlled environment for studying the effect of the tabular shield, we introduce a minimal domain that has the essential properties of our problem: a variety of (rare) events that can lead to catastrophic actions, a clear safety criterion, and a discrete state space. %
In this domain, agents are faced with a sequence of environments (POMDP-CAs), where the environment of each episode is drawn from a long-tailed distribution as illustrated in Figure~\ref{fig:LavaGrid}(left).
The domain we use for generating a sequence of POMDP-CAs, called \emph{LavaGrid}, is a grid domain based on the MiniGrid environment~\citep{gym_minigrid}. 

In LavaGrid, an agent faces a series of navigation tasks in which it has to reach a (green) goal position from its (red) starting position, while partially observing the environment. The agent can observe all tiles located within a $5 \times 5$ window from its current position in the direction it is facing, as well as the relative position of the goal in each dimension.
The start and goal position are fixed between the different tasks, as is the layout of the map (number of tiles and position of walls). However, in some tasks, tiles could be covered with \emph{lava}. If the agent steps into a tile with lava, the episode ends, and the agent receives a negative reward. At the beginning of each episode, a new environment is generated in which each tile is covered by lava with some fixed probability (independent of other tiles); the probability of a tile being covered with lava exponentially increases along the optimal path to the goal. The structure of the map and the probability of each tile to be covered with lava are shown in Figure~\ref{fig:LavaGrid}(right). In addition, whenever a tile with lava is generated, it is assigned one of three possible types. Red lava is the most common (94\% probability), blue lava is much scarcer (5\% probability), and purple lava is the rarest type (1\% probability).

The above schema creates the desired long-tailed distribution of tasks as shown in Figure~\ref{fig:LavaGrid}(left). Usually (with probability $\sim 94\%$) there is no lava at all and the environment is safe, but there are a total of $2 \times 10^{25}$ possible lava configurations which the agent can encounter, where each configuration corresponds to a different POMDP-CA. Importantly, each lava configuration is meant to represent a different class of events, i.e., stepping into lava placed on one tile is an entirely different mistake from stepping into lava on another tile. Thus, generalization cannot occur between different classes of events. For example, consider red lava on one square as representing a dog running in front of a car, and red lava on a neighboring square as an ice patch in front of the car. In order to prevent generalization between different classes of events, agents cannot directly observe lava. Rather, they receive a one-hot vector which indicates which POMDP-CA they are currently facing. Since there are too many POMDP-CAs to encode in a one-hot vector representation, all POMDP-CAs that corresponds to low probability (less than $2\cdot 10^{-8}$) lava configurations are clustered together to a single entry in the vector.

\subsection{Experimental Results}
In all experiments, we report the mean results of five different runs on the LavaGrid domain, where the reward of reaching the goal was set to $10$, the reward of stepping into lava was set to $-1000$, and the reward from every non-terminal step was set to be the $L_1$ distance of the agent from the goal divided by the maximal distance. Each figure mentioned below is composed of two parts. The left side shows the mean episodic return (on a symmetrical log scale), while the right side shows the mean rate of mistakes (total mistake divided by the number of steps, on a log scale). The standard error of the different runs is shown as a shaded area around each line. The WCSAC algorithm is executed using two different CVaR thresholds, used for controlling the tradeoff between risk and reward: a value of 0.25 which encourages risk averseness, and a value of 0.75 which generally results in polices that take more risks.

\paragraph{Single-agent Experiments}
\begin{figure*}[bth]
     \centering
         \includegraphics[width=\textwidth]{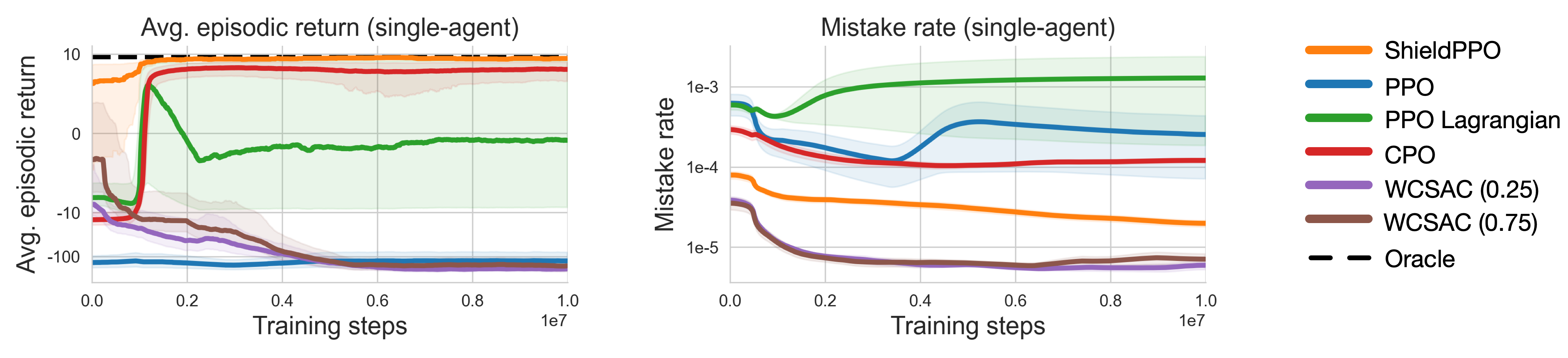}
         \caption{Single-agent results on the LavaGrid domain.}
         \label{fig:single_shield}
\end{figure*}

In the single-agent experiment, we expect to see that the mistake rate of ShieldPPO decreases overtime, as mistakes in the most-probable instances would be added to the table and would not be repeated. Also, ShieldPPO does not aim to avoid unsafe actions in general, but merely to not repeat actions which are known to be catastrophic. Thus, we expect ShieldPPO to achieve a high mean reward, while other safe RL algorithms might sacrifice performance in an attempt to act safely.

The results of this experiment are shown in Figure~\ref{fig:single_shield}. Here, in terms of mean return over time, ShieldPPO quickly converges to an optimal solution, and is the only algorithm to do so. CPO converges a bit more slowly to a good suboptimal policy. By contrast, all other safe RL baselines improve the average return in the first few episodes but end up converging to suboptimal policies that result in negative return, where the two instances of WCSAC converge to a policy that almost never moves in order to avoid risks. Interestingly, PPO, which is not concerned with safety, also mostly converges to a policy that avoids moving at all, tough in some runs it did manage converge to an optimal policy. We conjecture that since PPO repeats the same mistake more than once, and that mistakes are associated with highly negative rewards, PPO learns to avoid moving in an attempt to maximize the reward, while ShieldPPO learns to avoid repeating mistakes in the most probable instances quickly, and therefore can maximize reward while not fearing making mistakes often.
The results on the average mistake-rate show that ShieldPPO reduces its mistake rate over time, as expected, and has an overall low mistake rate. The only algorithms with lower mistake rate are the two WCSAC instances, in which the agent quickly learns to almost never move, thus avoiding mistakes almost completely. All algorithms other than ShieldPPO repeated at least one mistake within the first 200 episodes, while as expected, ShieldPPO never repeated a mistake.

\paragraph{Multi-agent Experiments}
\begin{figure*}[bth]
     \centering
         \includegraphics[width=\textwidth]{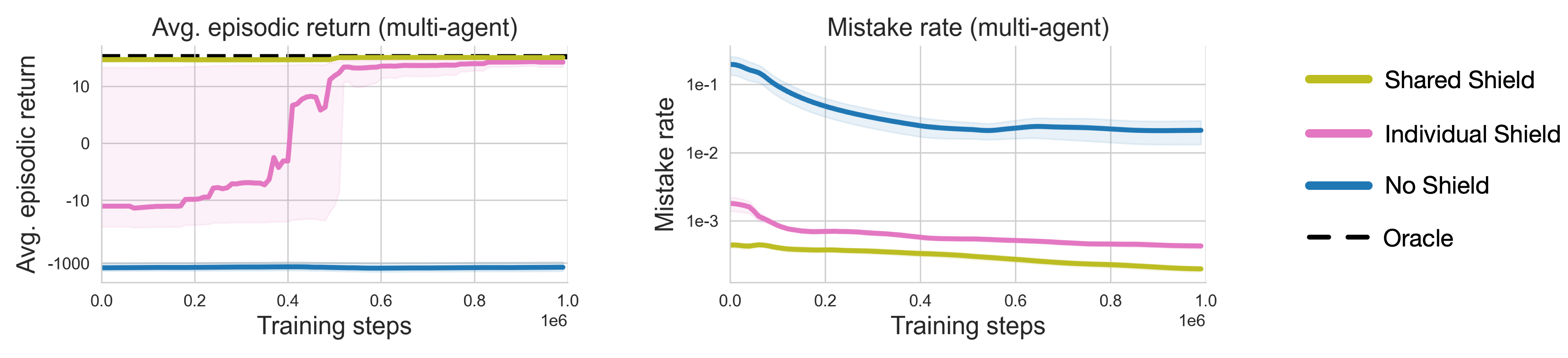}
         \caption{Multi-agent results on the LavaGrid domain}
         \label{fig:multi_shield}
\end{figure*}
In this experiment, we considered the sum of mean returns and the joint mistake-rate of ten different agents. We hypothesis that the shared-shield variant would have the lowest mistake rate and that it would be able to converge to a high mean reward faster than the other variants. This hypothesis is supported by the results reported in Figure~\ref{fig:multi_shield}, which show that the shared shield helps the agents to converge to the optimal policy $\sim\! 10$ times faster compared to individual shields, and to preform fewer mistakes as a group. When a shield was not used, the agents were not able converge to the optimal policy, or even to a good suboptimal policy.

\paragraph{Goal-conditioned Experiments}

In this experiment, the goal of the agent can be one of three different locations (one of the three corners of the grids, other than the agent's starting position), and the probability of each tile to be covered lava is $0.5\%$ regardless of its position. In this setting, ShieldPPO is expected to have an even more pronounced advantage over the other algorithms, as the shield can be shared across different goal locations, while the other approaches cannot be easily share safety information between different goal locations (nor between different tasks). Indeed, the results show that all algorithms converge to a policy of not moving, except for ShieldPPO that was able to converge to the optimal policy after 30k steps on average (and 50k steps in the worst case).

\section{Conclusions and Future Research Directions}
This paper takes a first step towards enabling learning agents to avoid mistake repetition by learning a shield online based on catastrophic action effects that agents experience. The empirical results indicate that this approach has a great potential to improve the performance and safety of agents. Nonetheless, there are three strong assumptions that needs to be relaxed for this approach to be useful in real-life scenarios: discrete environment, sufficient memory, and the ability to identify mistakes (to observe $L_\phi$). One possible future research direction to address the first two assumptions is finding an effective design for hybrid shields that can generalize across different mistakes in continuous domains, while using a fixed memory. Another potential research direction is relaxing the third assumption and assuming that $L_\phi$ is observable only approximately, e.g., only being able to know that a trajectory in which the car ended up hitting a wall contains a mistake, but not exactly when this mistake occurred. One promising approach to the latter is to estimate the conditional Value-at-Risk (CVaR) of actions to assess their safety, and to treat actions which are deemed to be risky as mistakes.

\section*{Acknowledgements}
This work has taken place in the Learning Agents Research
Group (LARG) at the Artificial Intelligence Laboratory, The University
of Texas at Austin.  LARG research is supported in part by the
National Science Foundation (CPS-1739964, IIS-1724157, FAIN-2019844),
the Office of Naval Research (N00014-18-2243), Army Research Office
(W911NF-19-2-0333), DARPA, Lockheed Martin, General Motors, Bosch, and
Good Systems, a research grand challenge at the University of Texas at
Austin.  The views and conclusions contained in this document are
those of the authors alone.  Peter Stone serves as the Executive
Director of Sony AI America and receives financial compensation for
this work.  The terms of this arrangement have been reviewed and
approved by the University of Texas at Austin in accordance with its
policy on objectivity in research.

\bibliography{reference}

\end{document}